\pdfoutput=1
\documentclass[11pt]{article}
\usepackage{naacl2021}
\usepackage{times,booktabs, multirow, graphicx, endnotes, hyperref}
\usepackage{latexsym}
\usepackage{rotating}
\usepackage{amssymb, amsmath}
\usepackage[T1]{fontenc}
\usepackage[utf8]{inputenc}
\usepackage{microtype}
\usepackage[normalem]{ulem}
\useunder{\uline}{\ul}{}

\newcommand{\ie}{{\it i.e.}}
\newcommand{\eg}{{\it e.g.}}
\newcommand{\com}{\textcolor{red}}

\title{MelBERT: Metaphor Detection via Contextualized Late Interaction using Metaphorical Identification Theories}

\author{Minjin Choi$^1$, Sunkyung Lee$^1$, Eunseong Choi$^1$, Heesoo Park$^2$, \\ \textbf{Junhyuk Lee$^1$, Dongwon Lee$^3$} and \textbf{Jongwuk Lee$^1$}\\
  $^1$Sungkyunkwan University, Republic of Korea, $^2$Bering Lab, Republic of Korea, \\
  $^3$The Pennsylvania State University, United States\\
  \texttt{$^1$\{zxcvxd,sk1027,eunseong,ljhjoon00,jongwuklee\}@skku.edu} \\
  \texttt{$^2$heesoo.park@beringlab.com} \texttt{$^3$dongwon@psu.edu} \\}

\date{}

\begin{document}
\maketitle

\begin{abstract}
Automated metaphor detection is a challenging task to identify the metaphorical expression of words in a sentence. To tackle this problem, we adopt pre-trained contextualized models, \eg, BERT and RoBERTa. To this end, we propose a novel metaphor detection model, namely \emph{metaphor-aware late interaction over BERT (MelBERT)}. Our model not only leverages contextualized word representation but also benefits from linguistic metaphor identification theories to detect whether the target word is metaphorical. Our empirical results demonstrate that MelBERT outperforms several strong baselines on four benchmark datasets, \ie, VUA-18, VUA-20, MOH-X, and TroFi.
\end{abstract}

\section{Introduction}

As the conceptual and cognitive mapping of words, a \emph{metaphor} is a common language expression representing other concepts rather than taking literal meanings of words in context~\cite{LakoffJ80, LagerwerfM08}. For instance, in the sentence ``hope is on the horizon,'' the word ``horizon'' does not literally mean the line at the earth's surface. It is a metaphorical expression to describe a positive situation. Therefore, the meaning of ``horizon'' is context-specific and different from its literal definition.

As the metaphor plays a key role in cognitive and communicative functions, it is essential to understand contextualized and unusual meanings of words (\eg, metaphor, metonymy, and personification) in various natural language processing (NLP) tasks, \eg, machine translation~\cite{ShiIL14}, sentiment analysis~\cite{CambriaPGT17}, and dialogue systems~\cite{DybalaS12}. A lot of existing studies have developed various computational models to recognize metaphorical words in a sentence.

Automated metaphor detection aims at identifying metaphorical expressions using computational models. Existing studies can be categorized into three pillars. First, feature-based models employ various hand-crafted features~\cite{ShutovaSK10, TurneyNAC11, ShutovaS13, BroadwellBCSFTSLCW13, TsvetkovBGND14, BulatCS17}. Although simple and intuitive, they are highly sensitive to the quality of a corpus. Second, some studies~\cite{WuWCWYH18, GaoCCZ18, MaoLG19} utilize recurrent neural networks (RNNs), which are suitable for analyzing the sequential structure of words. However, they are limited to understanding the diverse meanings of words in context. Lastly, the pre-trained contextualized models, \eg, BERT~\cite{DevlinCLT19} and RoBERTa~\cite{LiuOGDJCLLZS19}, have been used for detecting metaphors~\cite{ChenLFK20, GongGJB20, SuFHLWC20}. Owing to the powerful representation capacity, such models have been successful for addressing various NLP tasks~\cite{WangSMHLB19} and document ranking in IR~\cite{MitraC18}.

Based on such an advancement, we utilize a contextualized model using two metaphor identification theories, \ie, \emph{Metaphor Identification Procedure (MIP)}~\cite{Group07, SteenDHKKP10} and \emph{Selectional Preference Violation (SPV)}~\cite{Wilks75, Wilks78}. For MIP, a metaphorical word is recognized if the literal meaning of a word is different from its contextual meaning~\cite{haagsma2016}. For instance, in the sentence ``Don't twist my words'', the contextual meaning of ``twist'' is ``to distort the intended meaning'', different from its literal meaning, ``to form into a bent, curling, or distorted shape.'' For SPV, a metaphorical word is identified if the target word is unusual in the context of its surrounding words. That is, ``twist'' is metaphorical because it is unusual in the context of ``words.'' Although the key ideas of the two strategies are similar, they have different procedures for detecting metaphorical words and their contexts in the sentence.

To this end, we propose a novel metaphor detection model using metaphorical identification theories over the pre-trained contextualized model, namely \emph{metaphor-aware late interaction over BERT} (\emph{MelBERT}). MelBERT deals with a classification task to identify whether a target word in a sentence is metaphorical or not. As depicted in Figure~\ref{fig:melbert}, MelBERT is based on a siamese architecture that takes two sentences as input. The first sentence is a sentence $S$ with a target word $w_t$ and the second sentence is a target word $w_t$ itself. MelBERT independently encodes $S$ and $w_t$ into each embedding vector, which avoids unnecessary interactions between $S$ and $w_t$. Inspired by MIP, MelBERT then employs the contextualized and isolated representations of $w_t$ to distinguish between the contextual and literal meaning of $w_t$. To utilize SPV, MelBERT employs the sentence embedding vector and the contextualized target word embedding vector. MelBERT identifies how much the surrounding words mismatch from the target word. Lastly, MelBERT combines two metaphor identification strategies to predict if a target word is metaphorical or not. Each metaphor identification theory is non-trivial for capturing complicated and vague metaphorical words. To overcome these limitations, we incorporate two linguistic theories into a pre-trained contextualized model and utilize several linguistic features such as POS features.

To summarize, MelBERT has two key advantages. First, MelBERT effectively employs the contextualized representation to understand various aspects of words in context. Because MelBERT is particularly based on a late interaction over contextualized models, it can prevent unnecessary interactions between two inputs and effectively distinguish the contextualized meaning and the isolated meaning of a word. Second, MelBERT utilizes two metaphor identification theories to detect whether the target word is metaphorical. Experimental results show that MelBERT consistently outperforms state-of-the-art metaphor detection models in terms of F1-score on several benchmark datasets, such as VUA-18, VUA-20, and VUA-Verb datasets.

\section{Related Work}

\subsection{Metaphor Detection}

\noindent
\textbf{Feature-based approach}. Various linguistic features are used to understand metaphorical expressions. Representative hand-engineered features include word abstractness and concreteness~\cite{TurneyNAC11}, word imageability~\cite{BroadwellBCSFTSLCW13},  semantic supersenses~\cite{TsvetkovBGND14}, and property norms~\cite{BulatCS17}. However, they have difficulties handling rare usages of metaphors because the features rely on manually annotated resources. To address this problem, sparse distributional features~\cite{ShutovaSK10, ShutovaS13} and dense word embeddings~\cite{ShutovaKM16, ReiBKS17}, \ie, Word2Vec~\cite{MikolovSCCD13}, are used as better linguistic features. For details, refer to the survey~\cite{VealeSK16}.

\vspace{1mm}
\noindent
\textbf{RNN-based approach}. Several studies proposed neural metaphor detection models using recurrent neural networks (RNNs). \cite{WuWCWYH18} adopts a bidirectional-LSTM (BiLSTM)~\cite{GravesS05} and a convolutional neural network (CNN) using Word2Vec~\cite{MikolovSCCD13} as text features in addition to part-of-speech (POS) and word clustering information as linguistic features. ~\cite{GaoCCZ18} employs BiLSTM as an encoder using GloVe~\cite{PenningtonSM14} and ELMo~\cite{PetersNIGCLZ18} as text input representation. ~\cite{MaoLG19} makes use of the metaphor identification theory on top of the architecture of ~\cite{GaoCCZ18}. Despite their success, the shallow neural networks (\eg, BiLSTM and CNN) have limitations on representing various aspects of words in context.

\vspace{1mm}
\noindent
\textbf{Contextualization-based approach}. Recent studies utilize pre-trained contextualized language models, \eg, BERT~\cite{DevlinCLT19} and RoBERTa~\cite{LiuOGDJCLLZS19}, for metaphor detection. Because the pre-trained model can encode rich semantic and contextual information, it is useful for detecting metaphors with fine-tuning training. DeepMet~\cite{SuFHLWC20} utilizes RoBERTa with various linguistic features, \ie, global text context, local text context, and POS features. IlliniMet~\cite{GongGJB20} combines RoBERTa with linguistic information obtained from external resources. ~\cite{ChenLFK20} formulates the multi-task learning problem for both metaphor detection, and ~\cite{LeongKHSUC20} reports the results of these models in the VUA 2020 shared task.

\subsection{Semantic Matching over BERT}

The key idea of neural semantic matching is that neural models encode a query-document pair into two embedding vectors and compute a relevance score between the query and the document~\cite{MitraC18}. The simple approach is to feed a query-document pair to BERT~\cite{DevlinCLT19} and compute a relevance score, where the query and the document are fully interacted~\cite{NogueiraYLC19, DaiC20}. In contrast, SBERT~\cite{ReimersG19}, TwinBERT~\cite{LuJZ20}, and ColBERT~\cite{KhattabZ20} adopt late interaction architectures using siamese BERT, where the query and the document are encoded independently. Our work is based on the late interaction architecture. In other words, the sentence with the target word and the target word is encoded separately to represent contextualized and isolated meanings of the target word.
\section{MelBERT}

In this section, we propose a novel metaphor detection model over a pre-trained contextualized model. To design our model, we consider two metaphor detection tasks. Given a sentence $S = \{ w_1, \dots, w_{n}\}$ with $n$ words and a target word $w_t \in S$, the classification task predicts the metaphoricity (\ie, mataphorical or literal) of $w_t$. Given a sentence $S$, the sequence labeling predicts the metaphoricity of each word $w_t$ ($1 \le t \le n$) in $S$.

We aim at developing a metaphor detection model for the classification task. Our model returns a binary output, \ie, 1 if the target word $w_t$ in $S$ is metaphorical or 0 otherwise. By sequentially changing the target word $w_t$, our model can be generalized to classify the metaphoricity of each word in a sentence, as in sequence labeling.

\subsection{Motivation}

The pre-trained language models, \eg, BERT~\cite{DevlinCLT19} and RoBERTa~\cite{LiuOGDJCLLZS19}, usually take two sentences as input and return output to predict the relevance between two input sentences. We adopt RoBERTa as the contextualized backbone model because RoBERTa is known to outperform BERT~\cite{LiuOGDJCLLZS19}. To design a metaphor detection model, we treat one input sentence as a single word (or a phrase).

\begin{figure}
\centering
\begin{small}
\begin{tabular}{cc}
\includegraphics[width=0.21\textwidth]{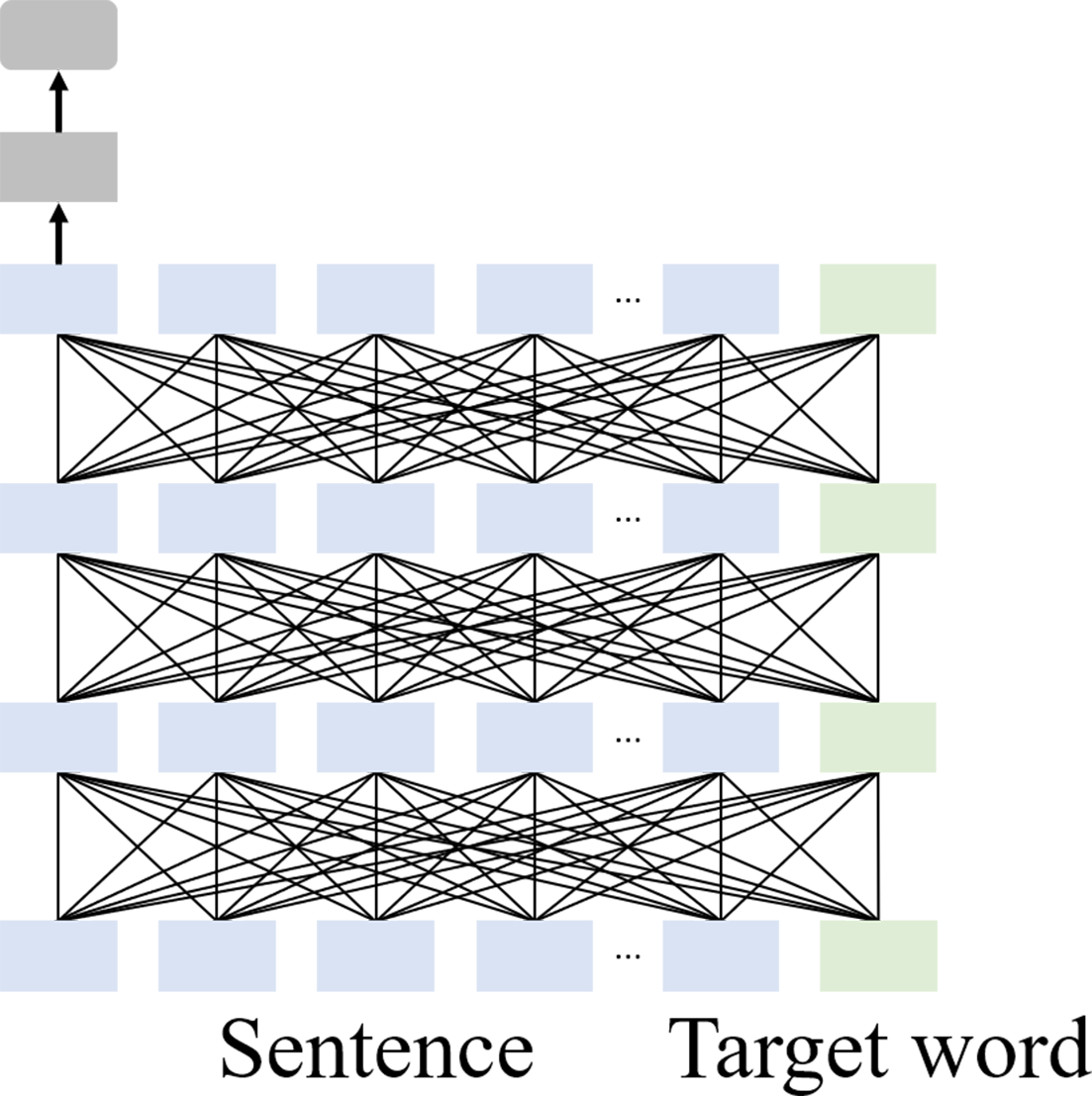} & 
\includegraphics[width=0.21\textwidth]{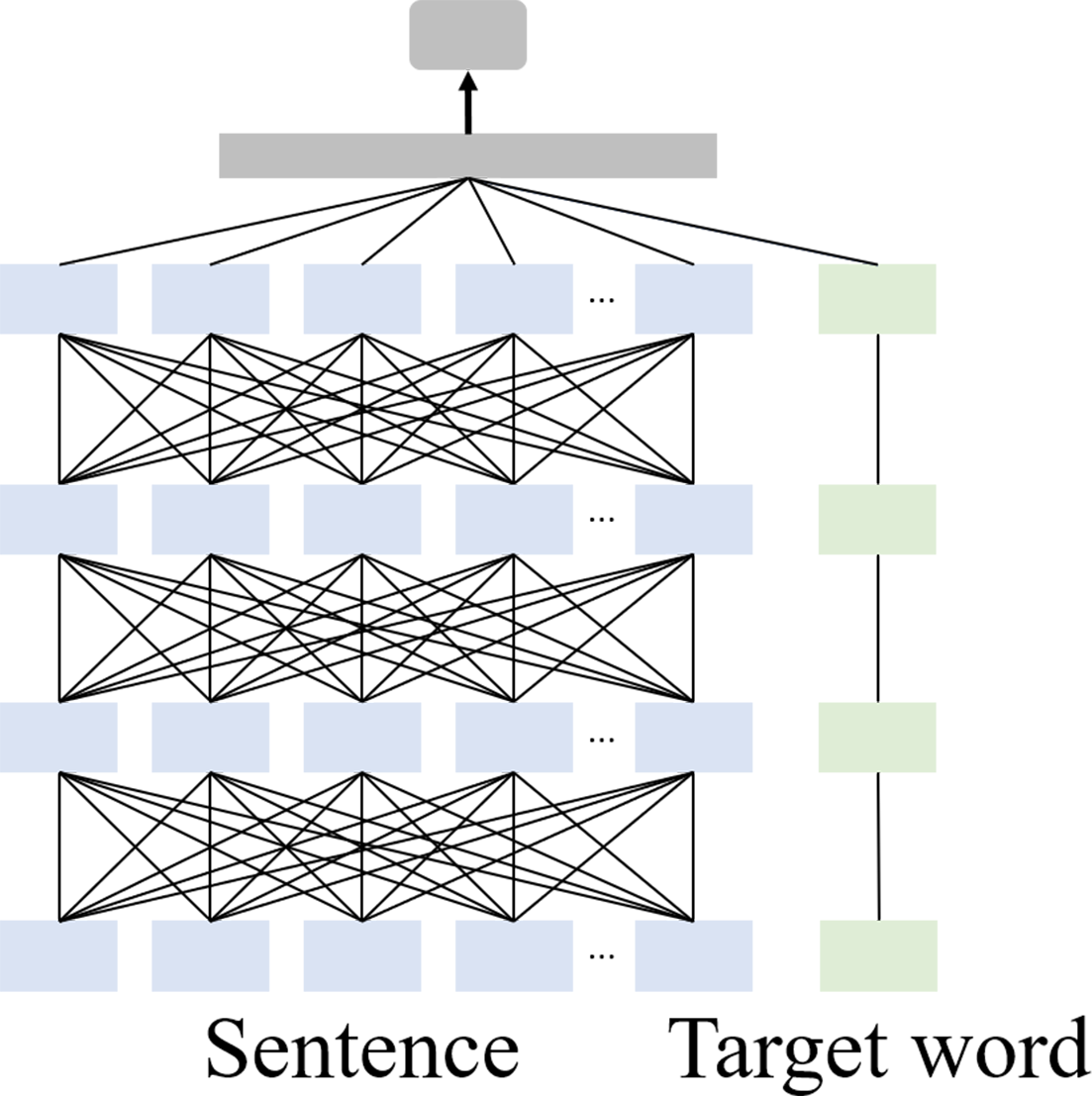} \\
(a) All-to-all interaction & (b) Late interaction \\
\end{tabular}
\end{small}
\vskip -0.1in
\caption{Two interaction paradigms over a contextualized model.}
\label{fig:two_paradigms}
\vskip -0.1in
\end{figure}

As depicted in Figure~\ref{fig:two_paradigms}, there are two paradigms for representing the interaction between two input sentences: \emph{all-to-all interaction} and \emph{late interaction}, as discussed in the document ranking problem~\cite{KhattabZ20}. While all-to-all interaction takes two input sentences together as an input, late interaction encodes two sentences separately over a siamese architecture. Given a sentence $S$ and a target word $w_t$, all-to-all interaction can capture all possible interactions within and across $w_t$ and $S$, which incurs high computational cost. Moreover, when some interactions across $w_t$ and $S$ are useless, it may learn noisy information. In contrast, because late interaction encodes $w_t$ and $S$ independently, it naturally avoids unnecessary intervention across $w_t$ and $S$. The sentence embedding vector also can be easily reused in computing the interaction with the target word. In other words, the cost of encoding the sentence vector can be amortized for that of encoding different target words.

Because our goal is to identify whether the contextualized meaning of the target word $w_t$ is different from its isolated meaning, we adopt the late interaction paradigm for metaphor detection. Our model encodes a sentence $S$ with a target word and a target word $w_t$ into embedding vectors, respectively, and computes the metaphoricity score of the target word. (In Section~\ref{sec:experiments}, it is found that our model using late interaction outperforms a baseline model using all-to-all interaction.)

\subsection{Model Architecture}

\begin{figure*}[t]
    \centering
        \includegraphics[width=1.0\textwidth]{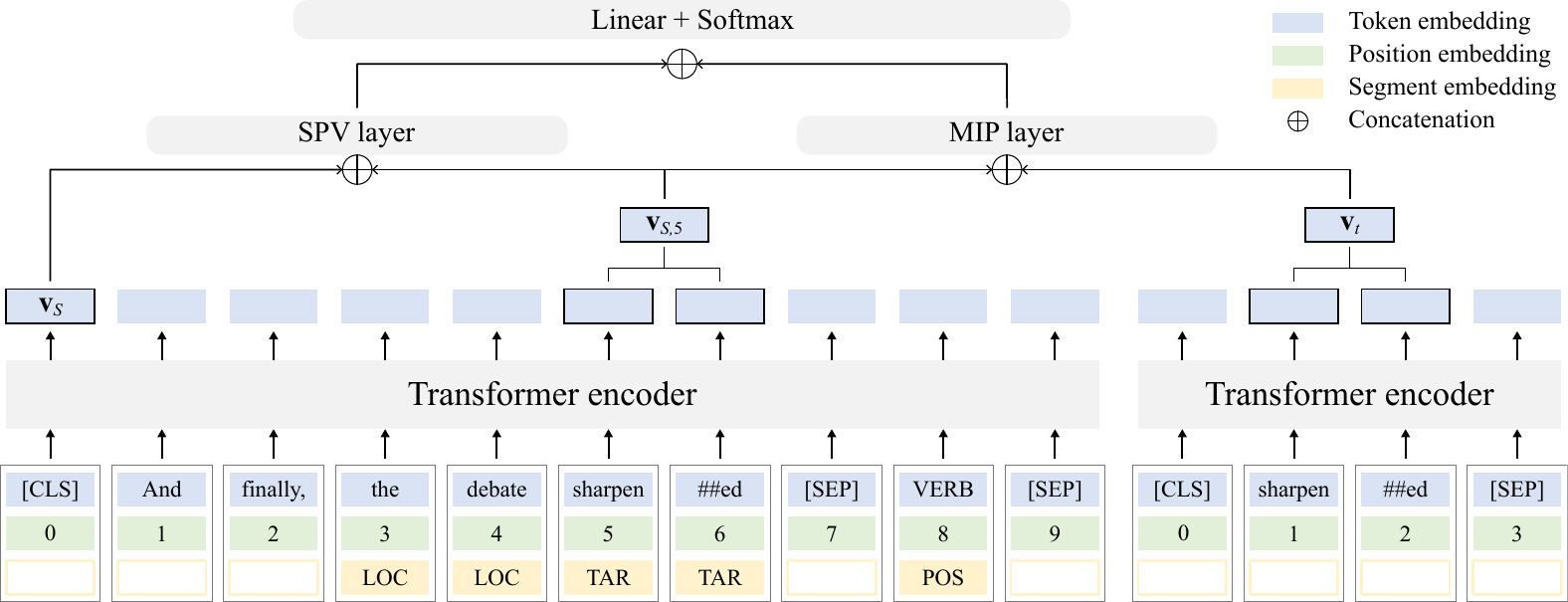}
    \caption{Model architecture of MelBERT. When a target word $w_t$ is split into multiple tokens by BPE, the average pooling is used for the target word.}\label{fig:melbert} \vspace{-1mm}
\end{figure*}

We propose a novel metaphor detection model, namely, \emph{metaphor-aware late interaction over BERT} (\emph{MelBERT}) using metaphor identification theories, \ie, \emph{Metaphor Identification Procedure (MIP)}~\cite{Group07, SteenDHKKP10} and \emph{Selectional Preference Violation (SPV)}~\cite{Wilks75, Wilks78}. Figure~\ref{fig:melbert} illustrates the overall architecture of MelBERT, which consists of three components: a sentence encoder $Enc(S)$, a target word encoder $Enc(w_t)$, and a late interaction mechanism to compute a score.

We first explain the input layer for two encoders $Enc(S)$ and $Enc(w_t)$. Each word in the sentence is converted to tokens using an improved implementation of byte-pair encoding (BPE)~\cite{RadfordWCLAS19}. As shown in the original BERT, the position embedding is used to represent the position of tokens. The segment embedding is used to distinguish target tokens (denoted as [TAR]) and their local context (denoted as [LOC]). When the sentence is represented as a composite sentence, the local context indicates a clause including target tokens. For simplicity, we represent the local context using comma separator (,) in the sentence. Besides, we add a special classification token [CLS] before the first token and a segment separation token [SEP] after the last token. To make use of the POS feature of the target word, we append the POS tag for the target word after [SEP], as used in~\cite{SuFHLWC20}. The input representation is finally computed by the element-wise addition of token, position embedding, and segment embedding. For $Enc(w_t)$, the target word is converted to the tokens using BPE, but position and segment embedding are not used.

Given a sentence $S = \{w_1, \dots, w_{n}\}$, $Enc(S)$ encodes each word into a set of contextualized embedding vectors, $\{\mathbf{v}_{S}, \mathbf{v}_{S, 1}, \ldots, \mathbf{v}_{S, n}\}$ using the transformer encoder~\cite{VaswaniSPUJGKP17}, where $\mathbf{v}_{S}$ is the embedding vector corresponding to the [CLS] token and $\mathbf{v}_{S, i}$ is the $i$-th embedding vector for $w_i$ in $S$. Similarly, $Enc(w_t)$ encodes a target word $w_t$ into $\mathbf{v}_{t}$ without context.
\begin{equation}\small
\begin{split}
\mathbf{v}_{S}, \mathbf{v}_{S, 1}, \ldots, \mathbf{v}_{S, n} = & \\
  Enc(``& [CLS], w_1, \ldots, w_{n}, [SEP]")
\end{split}
\end{equation}
\begin{equation}\small
\mathbf{v}_t = Enc(``[CLS], w_t, [SEP]")
\end{equation}

While $\mathbf{v}_S$ reflects the interaction across all words in $S$, $\mathbf{v}_{S, t}$ considers the interaction between $w_t$ and other words in $S$. Therefore, $\mathbf{v}_{S, t}$ and $\mathbf{v}_t$ can be interpreted as different meanings for $w_t$, \ie, $\mathbf{v}_{S, t}$ is contextualized representation of $w_t$ and $\mathbf{v}_t$ is isolated representation of $w_t$.

Then, we utilize two metaphor identification theories using contextualized embedding vectors.

\vspace{1mm}
\noindent
\textbf{MelBERT using MIP}. The basic idea of MIP is that a metaphorical word is identified by the gap between the contextual and literal meaning of a word. To incorporate MIP into MelBERT, we employ two embedding vectors $\mathbf{v}_{S, t}$ and $\mathbf{v}_t$, representing a contextualized embedding vector and an isolated embedding vector for $w_t$, respectively. Using these vectors, we identify the semantic gap for the target word in context and isolation.

\vspace{1mm}
\noindent
\textbf{MelBERT using SPV}. The idea of SPV is that a metaphorical word is identified by the semantic difference from its surrounding words. Unlike MIP, we only utilize the sentence encoder. Given a target word $w_t$ in $S$, our key assumption is that $\mathbf{v}_{S}$ and $\mathbf{v}_{S, t}$ show a semantic gap if $w_t$ is metaphorical. Although $\mathbf{v}_{S}$ and $\mathbf{v}_{S, t}$ are contextualized, the meanings of the two vectors are different; $\mathbf{v}_{S}$ represents the interaction across all pair-wise words in $S$, but $\mathbf{v}_{S, t}$ represents the interaction between $w_t$ and other words in $S$. In this sense, when $w_t$ is metaphorical, $\mathbf{v}_{S, t}$ can be different from $\mathbf{v}_{S}$ by the surrounding words of $w_t$.

\vspace{1mm}
\noindent
\textbf{Late interaction over MelBERT}. Using the two strategies, MelBERT predicts whether a target word $w_t \in S$ is metaphorical or not. We can compute a hidden vector $h_{MIP}$ by concatenating $\mathbf{v}_{S, t}$ and $\mathbf{v}_t$ for MIP.
\begin{equation}\small
h_{MIP} = f([\mathbf{v}_{S, t}; \mathbf{v}_{t}]),
\end{equation}
\noindent
where $h_{MIP} \in \mathbb{R}^{h \times 1}$ and $f(\cdot)$ is a function for the MLP layer to learn the gap between two vectors $\mathbf{v}_{S, t}$ and $\mathbf{v}_{t}$.

We can also compute a hidden vector $h_{SPV}$ using $\mathbf{v}_{S}$ and $\mathbf{v}_{S, t}$ for SPV.
\begin{equation}\small
h_{SPV} = g([\mathbf{v}_{S}; \mathbf{v}_{S, t}]),
\end{equation}
\noindent
where $h_{SPV} \in \mathbb{R}^{h \times 1}$ and $g(\cdot)$ is a function for the MLP layer to learn the semantic difference between $\mathbf{v}_{S}$ and $\mathbf{v}_{S, t}$.

We combine two hidden vectors $h_{MIP}$ and $h_{SPV}$ to compute a prediction score:
\begin{equation}\small
\hat{y} = \sigma(W^{\top} [h_{MIP}; h_{SPV}] + b),
\end{equation}

\noindent
where $\sigma(\cdot)$ is the sigmoid function, $W \in \mathbb{R}^{2h \times 1}$ is the parameter, and $b$ is a bias.
To learn MelBERT, finally, we use the cross-entropy loss function for binary classification as follows:
\begin{equation}\small
\mathcal{L} = \sum_{i=1}^{N}{y_i \log \hat{y}_i + (1 - y_i) \log (1 - \hat{y}_i)},    
\end{equation}
\noindent
where $N$ is the number of samples in the training set. $y_i$ and $\hat{y}_i$ are the true and predicted labels for the $i$-th sample in the training set.
\section{Evaluation}\label{sec:experiments}

In this section, we first present the experimental setup, then report empirical results by comparing our model against strong baselines.

\subsection{Experimental Setup}

\noindent
\textbf{Datasets}. We use four well-known public English datasets. First, the VU Amsterdam Metaphor Corpus (VUA) has been released in metaphor detection shared tasks in 2018 and 2020. We use two versions of VUA datasets, called VUA-18~\cite{LeongKS18} and VUA-20~\cite{LeongKHSUC20}, where VUA-20 is the extension of VUA-18. Let VUA-18$_{tr}$, VUA-18$_{dev}$, VUA-18$_{te}$ denote the training, validation, and test datasets, split from VUA-18. VUA-20$_{tr}$ includes VUA-18$_{tr}$ and VUA-18$_{dev}$. VUA-20$_{te}$ also includes VUA-18$_{te}$, and VUA-Verb$_{te}$ is a subset of VUA-18$_{te}$ and VUA-20$_{te}$.

Because most of the tokens in a sentence are literal words in VUA-18, VUA-20 selectively chooses the tokens in the training and testing datasets. VUA-18$_{te}$ consists of four genres, including news, academic, fiction, and conversation. It can also be categorized into different POS tags, such as verb, noun, adjective, and adverb. Additionally, we employ MOH-X~\cite{MohammadST16} and TroFi~\cite{BirkeS06} for testing purposes only. MOH-X is a verb metaphor detection dataset with the sentences from WordNet and TroFi is also a verb metaphor detection dataset, including sentences from the 1987-89 Wall Street Journal Corpus Release 1. The sizes of these datasets are relatively smaller than those of VUA datasets, and they have metaphorical words of more than 40\%, while VUA-18 and VUA-20 datasets have about 10\% of metaphorical words. While MOH-X and TroFi only annotate verbs as metaphorical words, the VUA dataset annotates all POS tags as metaphorical words. In this sense, we believe that the VUA dataset is more appropriate for training and testing models. Table~\ref{tab:dataset} summarizes detailed statistics on the benchmark datasets.

\begin{table}[t]
\begin{small}
\centering
\begin{tabular}{c|cccc}
\toprule
Dataset    & \multicolumn{1}{l}{\#tokens} & \multicolumn{1}{l}{\%M} & \multicolumn{1}{l}{\#Sent} & \multicolumn{1}{l}{Sent len} \\ \midrule
VUA-18$_{tr}$   & 116,622                      & 11.2                    & 6,323                     & 18.4                        \\
VUA-18$_{dev}$  & 38,628                       & 11.6                    & 1,550                     & 24.9                        \\
VUA-18$_{te}$   & 50,175                       & 12.4                    & 2,694                     & 18.6                        \\ \midrule
VUA-20$_{tr}$   & 160,154                      & 12.0                    & 12,109                    & 15                          \\
VUA-20$_{te}$   & 22,196                       & 17.9                    & 3,698                     & 15.5                        \\ \midrule
VUA-Verb$_{te}$ & 5,873                        & 30                      & 2,694                     & 18.6                        \\ \midrule
MOH-X      & 647                          & 48.7                    & 647                       & 8                           \\ \midrule
TroFi      & 3,737                        & 43.5                    & 3,737                     & 28.3                        \\ \bottomrule
\end{tabular}
\caption{Detailed statistics on benchmark datasets. \#tokens is the number of tokens, \%M is the percentage of metaphorical words, \#Sent is the number of sentences, and Sent len is the average length of sentences.}\label{tab:dataset}
\vspace{-1mm}
\end{small}
\end{table}

\vspace{1mm}
\noindent
\textbf{Baselines}. We compare our models with several strong baselines, including RNN-based and contextualization-based models. 
\begin{itemize}
    \item \textbf{RNN\_ELMo} and \textbf{RNN\_BERT}~\cite{GaoCCZ18}: They employ the concatenation of the pre-trained ELMo/BERT and the GloVe~\cite{PenningtonSM14} embedding vectors as an input, and use BiLSTM as a backbone model. Note that they use contextualized models only for input vector representation.

    \vspace{-1mm}
    \item \textbf{RNN\_HG} and \textbf{RNN\_MHCA}~\cite{MaoLG19}: They incorporate MIP and SPV into \textbf{RNN\_ELMo}~\cite{GaoCCZ18}. \textbf{RNN\_HG} compares an input embedding vector (literal) with its hidden state (contextual) through BiLSTM. \textbf{RNN\_MHCA} utilizes multi-head attention to capture the contextual feature within the window size. 
    
    \vspace{-1mm}
    \item \textbf{RoBERTa\_BASE}: It is a simple adoption of RoBERTa for metaphor detection. It takes a target word and a sentence as two input sentences and computes a prediction score. It can be viewed as a metaphor detection model over an all-to-all interaction architecture.
    
    \vspace{-1mm}
    \item \textbf{RoBERTa\_SEQ}~\cite{LeongKHSUC20}: It takes one single sentence as an input, and a target word is marked as the input embedding token and predicts the metaphoricity of the target word using the embedding vector of the target word. This architecture is used as the BERT-based baseline in the VUA 2020 shared task.

    \vspace{-1mm}
    \item \textbf{DeepMet}~\cite{SuFHLWC20}: It is the winning model in the VUA 2020 shared task. It also utilizes RoBERTa as a backbone model and incorporates it with various linguistic features, such as global context, local context, POS tags, and fine-grained POS tags.
\end{itemize}

\vspace{1mm}
\noindent
\textbf{Evaluation protocol}. Because the ratio of metaphorical words is relatively small, we adopt three metrics, \eg, precision, recall, and F1-score, denoted by Prec, Rec, and F1. MOH-X and TroFi datasets are too smaller than VUA datasets. Thus, we only used them as the test datasets; metaphor detection models are only trained in VUA datasets, and zero-shot transfer is conducted to evaluate the effectiveness of model generalization.

\vspace{1mm}
\noindent
\textbf{Implementation details}. For four baselines, we used the same hyperparameter settings\footnote{https://github.com/RuiMao1988/Sequential-Metaphor-Identification} in~\cite{GaoCCZ18, MaoLG19, SuFHLWC20}. For DeepMet\footnote{https://github.com/YU-NLPLab/DeepMet}, we evaluated it with/without bagging technique. While DeepMet~\cite{SuFHLWC20} exploits two optimization techniques, bagging and ensemble, we only used a bagging technique for MelBERT and DeepMet. It is because we want to evaluate the effectiveness of model designs. The performance difference for DeepMet between the original paper and ours thus comes from the usage of the ensemble method. For contextualized models, we used a pre-trained RoBERTa\footnote{https://huggingface.co/roberta-base} with 12 layers, 12 attention heads in each layer, and 768 dimensions of the hidden state. For contextualized baselines, we set the same hyperparameters with MelBERT, which were tuned on VUA-18$_{dev}$ based on F1-score. The batch size and max sequence length were set as 32 and 150. For training, the number of epochs was three with Adam optimizer. We increased the learning rate from 0 to 3e-5 during the first two epochs and then linearly decreased it during the last epoch. We set the dropout ratio as 0.2. All experimental results were averaged over five runs with different random seeds. We conducted all experiments on a desktop with 2 NVidia TITAN RTX, 256 GB memory, and 2 Intel Xeon Processor E5-2695 v4 (2.10 GHz, 45M cache). We implemented our model using PyTorch. All the source code is available at our website\footnote{https://github.com/jin530/MelBERT}.

\begin{table}[t]
\begin{small}
\begin{center}
\begin{tabular}{c|c|ccc}
\toprule
Dataset                    & Model         & Prec                   & Rec                  & F1                     \\ \midrule
\multirow{10}{*}{VUA-18}   & RNN\_ELMo     & 71.6                   & 73.6                   & 72.6                   \\
                           & RNN\_BERT     & 71.5                   & 71.9                   & 71.7                   \\
                           & RNN\_HG       & 71.8                   & {\ul \textit{76.3}}    & 74.0                   \\
                           & RNN\_MHCA     & 73.0                   & 75.7                   & 74.3                   \\ \cline{2-5} \rule{0pt}{1em}
                           & RoBERTa\_BASE & 79.4                   & 75.0                   & 77.1                   \\
                           & RoBERTa\_SEQ  & {\ul \textit{80.4}}    & 74.9                   & {\ul \textit{77.5}}    \\
                           & DeepMet       & \textbf{82.0}          & 71.3                   & 76.3                   \\
                           & MelBERT       & 80.1                   & \textbf{76.9}          & \textbf{78.5$^*$}          \\ \cline{2-5} \rule{0pt}{1em}
                           & DeepMet-CV  & 77.5                   & 80.2                   & 78.8                   \\
                           & MelBERT-CV  & \textit{\textbf{78.9}} & \textit{\textbf{80.7}} & \textit{\textbf{79.8$^*$}} \\ \midrule
\multirow{10}{*}{VUA-Verb} & RNN\_ELMo     & 68.2                   & 71.3                   & 69.7                   \\
                           & RNN\_BERT     & 66.7                   & 71.5                   & 69.0                   \\
                           & RNN\_HG       & 69.3                   & 72.3                   & 70.8                   \\
                           & RNN\_MHCA     & 66.3                   & \textbf{75.2}          & 70.5                   \\ \cline{2-5} \rule{0pt}{1em}
                           & RoBERTa\_BASE & 76.9                   & 72.8                   & 74.7                   \\
                           & RoBERTa\_SEQ  & {\ul \textit{79.2}}    & 69.8                   & 74.2                   \\
                           & DeepMet       & \textbf{79.5}          & 70.8                   & {\ul \textit{74.9}}    \\
                           & MelBERT       & 78.7                   & {\ul \textit{72.9}}    & \textbf{75.7}          \\ \cline{2-5} \rule{0pt}{1em}
                           & DeepMet-CV  & \textit{\textbf{76.2}} & 78.3                   & \textit{\textbf{77.2}} \\
                           & MelBERT-CV  & 75.5                   & \textit{\textbf{78.7}} & 77.1                   \\ \bottomrule
\end{tabular}
  \end{center}
  \vspace{-2mm}
  \caption{Performance comparison of MelBERT with baselines on VUA-18 and VUA-Verb (best is in \textbf{bold} and second best is in {\ul \textit{italic underlined}}). Let -CV denote the bagging technique for its base model (best is in \textit{\textbf{bold-italic}}). $*$ denotes $p < 0.05$ for a two-tailed t-test with the best competing model.}\label{tab:overall}
\end{small}\vspace{-1mm}
\end{table}%

\begin{table}[t]
\begin{small}
\begin{center}
\begin{tabular}{c|c|ccc}
\toprule
Dataset                 & Model         & Prec                  & Rec                   & F1                     \\ \midrule
\multirow{6}{*}{VUA-20} & RoBERTa\_BASE & 74.9                   & {\ul\textit{ 68.0}}             & 71.2                   \\
                        & RoBERTa\_SEQ  & \textbf{76.9}          & 66.7                   & {\ul \textit{71.4} }            \\
                        & DeepMet       & {\ul \textit{76.7}}             & 65.9                   & 70.9                   \\
                        & MelBERT       & 76.4                   & \textbf{68.6}          & \textbf{72.3$^*$}          \\ \cline{2-5}  \rule{0pt}{1em}
                        & DeepMet-CV    & 73.8                   & 73.2                   & 73.5                   \\
                        & MelBERT-CV    & \textit{\textbf{74.1}} & \textit{\textbf{73.7}} & \textit{\textbf{73.9}} \\ \bottomrule
\end{tabular}
  \end{center}
  \vspace{-2mm}
  \caption{Performance comparison of MelBERT with baselines on VUA-20 (best is in \textbf{bold} and second best is in {\ul \textit{italic underlined}}). Let -CV denote the bagging technique for its base model (best is in \textit{\textbf{bold-italic}}). $*$ denotes $p < 0.05$ for a two-tailed t-test with the best competing model.}\label{tab:overall_20}
\end{small}\vspace{-3mm}
\end{table}%

\subsection{Empirical Results}

\noindent
\textbf{Overall results}. Tables~\ref{tab:overall} and~\ref{tab:overall_20} report the comparison results of MelBERT against other baselines using RNNs and contextualized models on VUA-18, VUA-20, and VUA-Verb. It is found that MelBERT is consistently better than strong baselines in terms of F1-score. MelBERT outperforms (F1 = 78.5, 75.7, and 72.3) DeepMet~\cite{SuFHLWC20} with 2.8\%, 1.0\%, and 1.9\% performance gains on the three datasets. MelBERT also outperforms contextualized baseline models (\ie, RoBERTa\_BASE and RoBERTa\_SEQ), up to 1.2-1.5\% gains on the three datasets, indicating that MelBERT effectively utilizes metaphorical identification theories.

\begin{table}[t!]
\begin{small}
\begin{center}
\begin{tabular}{c|c|ccc}
\toprule
Genre                         & Model         & Prec               & Rec                & F1                  \\ \midrule
\multirow{8}{*}{Academic}   & RNN\_ELMo     & 78.2                & 80.2                & 79.2                \\
                              & RNN\_BERT     & 76.7                & 76.0                & 76.4                \\
                              & RNN\_HG       & 76.5                & \textbf{83.0}       & 79.6                \\
                              & RNN\_MHCA     & 79.6                & 80.0                & 79.8                \\ \cline{2-5} \rule{0pt}{1em}
                              & RoBERTa\_BASE & {\ul \textit{88.1}} & 79.5                & {\ul \textit{83.6}} \\
                              & RoBERTa\_SEQ  & 86.0                & 77.3                & 81.4                \\
                              & DeepMet       & \textbf{88.4}       & 74.7                & 81.0                \\ \cline{2-5} \rule{0pt}{1em}
                              & MelBERT       & 85.3                & {\ul \textit{82.5}} & \textbf{83.9}       \\ \midrule
\multirow{8}{*}{Conversation} & RNN\_ELMo     & 64.9                & 63.1                & 64.0                \\
                              & RNN\_BERT     & 64.7                & 64.2                & 64.4                \\
                              & RNN\_HG       & 63.6                & \textbf{72.5}       & 67.8                \\
                              & RNN\_MHCA     & 64.0                & 71.1                & 67.4                \\ \cline{2-5} \rule{0pt}{1em}
                              & RoBERTa\_BASE & 70.3                & 69.0                & 69.6                \\
                              & RoBERTa\_SEQ  & {\ul \textit{70.5}} & 69.8                & 70.1                \\
                              & DeepMet       & \textbf{71.6}       & 71.1                & \textbf{71.4}       \\ \cline{2-5} \rule{0pt}{1em}
                              & MelBERT       & 70.1                & {\ul \textit{71.7}} & {\ul \textit{70.9}} \\ \midrule
\multirow{8}{*}{Fiction}      & RNN\_ELMo     & 61.4                & 69.1                & 65.1                \\
                              & RNN\_BERT     & 66.5                & 68.6                & 67.5                \\
                              & RNN\_HG       & 61.8                & {\ul \textit{74.5}} & 67.5                \\
                              & RNN\_MHCA     & 64.8                & 70.9                & 67.7                \\ \cline{2-5} \rule{0pt}{1em}
                              & RoBERTa\_BASE & {\ul \textit{74.3}} & 72.1                & 73.2                \\
                              & RoBERTa\_SEQ  & 73.9                & 72.7                & {\ul \textit{73.3}} \\
                              & DeepMet       & \textbf{76.1}       & 70.1                & 73.0                \\ \cline{2-5} \rule{0pt}{1em}
                              & MelBERT       & 74.0                & \textbf{76.8}       & \textbf{75.4}       \\ \midrule
\multirow{8}{*}{News}         & RNN\_ELMo     & 72.7                & 71.2                & 71.9                \\
                              & RNN\_BERT     & 71.2                & 72.5                & 71.8                \\
                              & RNN\_HG       & 71.6                & \textbf{76.8}       & 74.1                \\
                              & RNN\_MHCA     & 74.8                & {\ul \textit{75.3}} & 75.0                \\ \cline{2-5} \rule{0pt}{1em}
                              & RoBERTa\_BASE & {\ul \textit{83.5}} & 71.8                & {\ul \textit{77.2}} \\
                              & RoBERTa\_SEQ  & 82.2                & 74.1                & \textbf{77.9}       \\
                              & DeepMet       & \textbf{84.1}       & 67.6                & 75.0                \\ \cline{2-5} \rule{0pt}{1em}
                              & MelBERT       & 81.0                & 73.7                & 77.2                \\ \bottomrule
\end{tabular}
 \end{center}
 \end{small}
 \vspace{-3mm}
 \caption{Model performance of different genres in VUA-18 (best is in \textbf{bold} and second best is in {\ul \textit{italic underlined}}).}\label{tab:genre}
 \vspace{-3mm}
\end{table}%


\begin{table}[t!]
\begin{small}
\begin{center}
\begin{tabular}{c|c|ccc}
\toprule
POS                        & Model         & Prec               & Rec                & F1                  \\ \midrule
\multirow{8}{*}{Verb}      & RNN\_ELMo     & 68.1                & 71.9                & 69.9                \\
                           & RNN\_BERT     & 67.1                & 72.1                & 69.5                \\
                           & RNN\_HG       & 66.4                & 75.5                & 70.7                \\
                           & RNN\_MHCA     & 66.0                & \textbf{76.0}       & 70.7                \\ \cline{2-5} \rule{0pt}{1em}
                           & RoBERTa\_BASE & {\ul \textit{77.0}} & 72.1                & 74.5                \\
                           & RoBERTa\_SEQ  & 74.4                & 75.1                & {\ul \textit{74.8}} \\
                           & DeepMet       & \textbf{78.8}       & 68.5                & 73.3                \\ \cline{2-5} \rule{0pt}{1em}
                           & MelBERT       & 74.2                & {\ul \textit{75.9}} & \textbf{75.1}       \\ \midrule
\multirow{8}{*}{Adjective} & RNN\_ELMo     & 56.1                & 60.6                & 58.3                \\
                           & RNN\_BERT     & 58.1                & 51.6                & 54.7                \\
                           & RNN\_HG       & 59.2                & \textbf{65.6}       & 62.2                \\
                           & RNN\_MHCA     & 61.4                & {\ul \textit{61.7}} & 61.6                \\ \cline{2-5} \rule{0pt}{1em}
                           & RoBERTa\_BASE & 71.7                & 59.0                & \textbf{64.7}       \\
                           & RoBERTa\_SEQ  & {\ul \textit{72.0}} & 57.1                & 63.7                \\
                           & DeepMet       & \textbf{79.0}       & 52.9                & 63.3                \\ \cline{2-5} \rule{0pt}{1em}
                           & MelBERT       & 69.4                & 60.1                & {\ul \textit{64.4}} \\ \midrule
\multirow{8}{*}{Adverb}    & RNN\_ELMo     & 67.2                & 53.7                & 59.7                \\
                           & RNN\_BERT     & 64.8                & 61.1                & 62.9                \\
                           & RNN\_HG       & 61.0                & 66.8                & 63.8                \\
                           & RNN\_MHCA     & 66.1                & 60.7                & 63.2                \\ \cline{2-5} \rule{0pt}{1em}
                           & RoBERTa\_BASE & 78.2                & {\ul \textit{69.3}} & {\ul \textit{73.5}} \\
                           & RoBERTa\_SEQ  & 77.6                & 63.9                & 70.1                \\
                           & DeepMet       & {\ul \textit{79.4}} & 66.4                & 72.3                \\ \cline{2-5} \rule{0pt}{1em}
                           & MelBERT       & \textbf{80.2}       & \textbf{69.7}       & \textbf{74.6}       \\ \midrule
\multirow{8}{*}{Noun}      & RNN\_ELMo     & 59.9                & 60.8                & 60.4                \\
                           & RNN\_BERT     & 63.3                & 56.8                & 59.9                \\
                           & RNN\_HG       & 60.3                & \textbf{66.8}       & 63.4                \\
                           & RNN\_MHCA     & 69.1                & 58.2                & 63.2                \\ \cline{2-5} \rule{0pt}{1em}
                           & RoBERTa\_BASE & \textbf{77.5}       & 60.4                & {\ul \textit{67.9}} \\
                           & RoBERTa\_SEQ  & 76.5                & 59.0                & 66.6                \\
                           & DeepMet       & {\ul \textit{76.5}} & 57.1                & 65.4                \\ \cline{2-5} \rule{0pt}{1em}
                           & MelBERT       & 75.4                & {\ul \textit{66.5}} & \textbf{70.7}       \\ \bottomrule
\end{tabular}
 \end{center}
 \end{small}
 \vspace{-3mm}
 \caption{Model performance of different POS tags in VUA-18 (best is in \textbf{bold} and second best is in {\ul \textit{italic underlined}}).}\label{tab:POS}
 \vspace{-3mm}
\end{table}

When combining MelBERT and DeepMet with the bagging technique, both models (\ie, MelBERT-CV and DeepMet-CV) show better performance than their original models by aggregating multiple models trained with 10-fold cross-validation process as used in~\cite{SuFHLWC20}. MelBERT-CV still shows better performance for all metrics than DeepMet-CV in VUA-18 and VUA-20. Also, MelBERT-CV (Recall = 73.7) significantly improves the original MelBERT (Recall = 68.6) in terms of recall. It implies that MelBERT-CV can capture various metaphorical expressions by combining multiple models.

Besides, it is found that contextualization-based models show better performance than RNN-based models in VUA-18 and VUA-Verb. While RNN-based models show 71-74\% F1-score, contextualization-based models show 76-78\% F1-score on VUA-18. It is revealed that RNN-based models are limited in capturing various aspects of words in context. Compared to RNN\_ELMo and RNN\_BERT, it also indicates that utilizing contextualization-based models as backbone models can have a better effect than simply utilizing it as an extra input embedding vector in~\cite{GaoCCZ18, MaoLG19}.

\vspace{1mm}
\noindent
\textbf{VUA-18 breakdown analysis}. Table~\ref{tab:genre} reports the comparison results for four genres in the VUA-18 dataset. MelBERT still shows better than or comparable to all competitive models in both breakdown datasets. Compared to RNN-based models, MelBERT achieves substantial improvements, as high as 4.9\% (Academic), 4.4\% (Conversation), 10.2\% (Fiction), and 2.8\% (News) in terms of F1-score. Particularly, they show the lowest accuracy because Conversation and Fiction have more complicated or rare expressions than other genres. For example, Conversation contains colloquial expressions or fragmented sentences such as ``ah'', ``cos'', ``yeah'' and Fiction often contains the names of fictional characters such as ``Tepilit'', ``Laibon'' which do not appear in other genres. Nonetheless, MelBERT shows comparable or the best performance in all genres. For Academic and Fiction, MelBERT particularly outperforms all the models in terms of F1-score.

Table~\ref{tab:POS} reports the comparison result for four POS tags in the VUA-18 dataset. For all POS tags, MelBERT consistently shows the best performance in terms of the F1-score. Compared to RNN-based models, MelBERT achieves as much as 5.9\% (Verb), 3.4\% (Adjective), 14.5\% (Adverb), and 10.3\% (Noun) gains in terms of F1-score. For all POS tags, MelBERT also outperforms DeepMet. It means that MelBERT using metaphorical identification theories can achieve consistent improvements regardless of POS tags of target words.

\begin{table}[t]
\begin{small}
\begin{center}
\begin{tabular}{c|c|ccc}
\toprule
Dataset                & Model         & Prec               & Rec                & F1                  \\ \midrule
\multirow{4}{*}{MOH-X} & RoBERTa\_BASE & 77.4                & \textbf{80.1}       & 78.4                \\
                       & RoBERTa\_SEQ  & \textbf{80.6}       & 77.7                & {\ul \textit{78.7}} \\
                       & DeepMet       & {\ul \textit{79.9}} & 76.5                & 77.9                \\ \cline{2-5} \rule{0pt}{1em} 
                       & MelBERT       & 79.3                & {\ul \textit{79.7}} & \textbf{79.2}       \\ \midrule
\multirow{4}{*}{TroFi} & RoBERTa\_BASE & \textbf{54.6}       & \textbf{74.3}       & \textbf{62.9}       \\
                       & RoBERTa\_SEQ  & 53.6                & 70.1                & 60.7                \\
                       & DeepMet       & {\ul \textit{53.7}} & 72.9                & 61.7                \\ \cline{2-5} \rule{0pt}{1em} 
                       & MelBERT       & 53.4                & {\ul \textit{74.1}} & {\ul \textit{62.0}} \\
\bottomrule
\end{tabular}
\end{center}
\end{small}
\vspace{-2mm}
\caption{Performance comparison of MelBERT with baselines over two datasets. Note that the models are trained on VUA-20, and these datasets are only used as the test datasets (best is in \textbf{bold} and second best is in {\ul \textit{italic underlined}}).}\label{tab:zero-shot}
\vspace{-2mm}
\end{table}

\begin{table}[t]
\begin{small}
\begin{center}
\begin{tabular}{c|ccc|ccc}
\toprule
\multirow{2}{*}{Model} & \multicolumn{3}{c|}{VUA-18}                   & \multicolumn{3}{c}{VUA-20}                    \\
                       & Prec         & Rec          & F1            & Prec         & Rec          & F1            \\ \midrule
MelBERT                & \textbf{80.1} & \textbf{76.9} & \textbf{78.5} & \textbf{76.4} & \textbf{68.6} & \textbf{72.3} \\ 
(-) MIP          & 77.8          & 75.8          & 76.7          & 74.7          & 67.8          & 71.1          \\
(-) SPV          & \textit{{\ul 79.5}}    & \textit{{\ul 76.3}}    & \textit{{\ul 77.9}}    & \textit{{\ul 74.9}}    & \textit{{\ul 68.6}}    & \textit{{\ul 71.7}}    \\
 \bottomrule
\end{tabular}
\end{center}
\end{small}
\vspace{-2mm}
\caption{Effect of different metaphorical identification theories on VUA-18 and VUA-20. (-) MIP and (-) SPV indicate MelBERT without MIP and SPV, respectively (best is in \textbf{bold} and second best is in {\ul \textit{italic underlined}}).}\label{tab:ablation_model}
\vspace{-3mm}
\end{table}


\vspace{1mm}
\noindent
\textbf{Zero-shot transfer on MOH-X and TroFi}. We evaluate a zero-shot learning transfer across different datasets, where the models are trained with the VUA-20 training dataset, and MOH-X and TroFi are used as test datasets. Although it is a challenging task, it is useful for evaluating the generalization power of trained models. Table~\ref{tab:zero-shot} reports the comparison results of MelBERT against other contextualization-based models. For the MOH-X dataset, MelBERT (F1 = 79.2) shows the best performance in terms of F1-score with 0.6--1.6\% performance gains. It indicates that MelBERT is an effective generalization model. For the TroFi dataset, the overall performance of all the models is much lower than MOH-X. It is because the average length of the sentences in the TroFi dataset is much longer and sentences are more complicated than those in MOH-X. Also, note that we trained DeepMet with the VUA-20 training dataset for evaluating a zero-shot transfer, while~\cite{SuFHLWC20} reported the results for DeepMet trained and tested with the MOH-X and TroFi datasets. While the performance gap between models is much small in terms of precision, MelBERT is better than DeepMet in terms of recall. It means that MelBERT can capture complicated metaphorical expressions than DeepMet.

\vspace{1mm}
\noindent
\textbf{Ablation study of MelBERT}. Table~\ref{tab:ablation_model} compares the effectiveness of metaphor identification theories. It is found that MelBERT using both strategies consistently shows the best performance. Also, MelBERT without SPV shows better performance than MelBERT without MIP, indicating that MelBERT using late interaction is more effective for capturing the difference between contextualized and isolated meanings of target words. Nonetheless, MelBERT shows the best performance by synergizing both metaphor identification strategies.

\begin{table}[t]
\begin{small}
\begin{center}
\begin{tabular}{ccc|c}
\toprule
\multicolumn{1}{l}{\multirow{4}{*}{\begin{turn}{90}(-) MIP\end{turn}}} & \multicolumn{1}{l}{\multirow{4}{*}{\begin{turn}{90}(-) SPV\end{turn}}} & \multicolumn{1}{l|}{\multirow{4}{*}{\begin{turn}{90}MelBERT\end{turn}}}      & \multirow{4}{*}{Sentence}    \\
\multicolumn{1}{l}{}                     & \multicolumn{1}{l}{}                     & \multicolumn{1}{l|}{}                           &                              \\ 
\multicolumn{1}{l}{}                     & \multicolumn{1}{l}{}                     & \multicolumn{1}{l|}{}                           &                              \\ 
\multicolumn{1}{l}{}                     & \multicolumn{1}{l}{}                     & \multicolumn{1}{l|}{}                           &                              \\ \midrule
\checkmark      &               & \checkmark     & Manchester is not \textit{\com{alone}}.  \\ \midrule
\checkmark      &               & \checkmark     & That's an \textit{\com{old}} trick.  \\ \midrule
\checkmark      &               & \checkmark     & Oh you rotten old \textit{\com{pig}}, you've been sick.  \\ \midrule
 \checkmark      &               & \checkmark     & "Are the twins \textit{\com{trash}}?"   \\ \midrule
                & \checkmark    & \checkmark     & I know, what is \textit{\com{going}} on! \\ \midrule
                & \checkmark    & \checkmark     & So who's \textit{\com{covering}} tomorrow? \\ \midrule
                & \checkmark    & \checkmark     & Do you \textit{\com{feel}} better now? \\ \midrule
                 & \checkmark    & \checkmark     & The day thrift turned into a \textit{\com{nightmare}}. \\ \midrule
                &               &                & \textit{\com{Way}} of the \textit{\com{World}}: Farming notes \\ \midrule
                &               &                & So many places Barry are \textit{\com{going}} down \\ \midrule
                &               &                & \textit{\com{Sensitivity}}, though, is not enough. \\ \bottomrule
\end{tabular}
\end{center}
\end{small}
\vspace{-2mm}
\caption{Examples of incorrect samples for MelBERT on VUA-20. The metaphorical words in the sentence are in red italicized. \checkmark marks correct model prediction.}\label{tab:errorcase}
\vspace{-3mm}
\end{table}

\vspace{1mm}
\noindent
\textbf{Error analysis}. Table~\ref{tab:errorcase} reports qualitative evaluation results of MelBERT. Based on the original annotation guideline\footnote{http://www.vismet.org/metcor/documentation/home.html}, we analyze several failure cases of MelBERT. For MelBERT without MIP, it is difficult to find common words with multiple meanings, \eg, \emph{go} and \emph{feel}. Also, when a sentence includes multiple metaphorical words, it mostly fails to detect metaphorical words. In this case, the surrounding words of a target word are not a cue to detect metaphors using SPV. Meanwhile, MelBERT without SPV has a failure case if target words are metaphorical for personification. That is, using MIP only, the target word can be closely interpreted by its literal meaning. As the most difficult case, MelBERT often fails to identify metaphorical words for borderline or implicit metaphors, \eg, \textit{``Way of the World''} is poetic.


\section{Conclusion}

In this work, we proposed a novel metaphor detection model, namely, \emph{metaphor-aware late interaction over BERT} (\emph{MelBERT}), marrying pre-trained contextualized models with metaphor identification theories. To our best knowledge, this is the first work that takes full advantage of both contextualized models and metaphor identification theories. Comprehensive experimental results demonstrated that MelBERT achieves state-of-the-art performance on several datasets.

\section*{Acknowledgments}

This work was supported by the National Research Foundation of Korea (NRF) (NRF-2018R1A5A1060031). Also, this work was supported by the Institute of Information \& communications Technology Planning \& evaluation (IITP) grant funded by the Korea government (MSIT) (No.2019-0-00421, AI Graduate School Support Program and No.2019-0-01590, High-Potential Individuals Global Training Program). The work of Dongwon Lee was in part supported by NSF awards \#1742702, \#1820609, \#1909702, \#1915801, and \#1934782.

\bibliographystyle{acl_natbib}
\bibliography{references}


\end{document}